\documentclass[preprint,review,12pt]{elsarticle}

\usepackage{graphicx}
\usepackage{amsmath,amsfonts,amssymb}
\usepackage{float}
\usepackage[table,x11names]{xcolor}
\usepackage{multirow}
\usepackage[utf8]{inputenc}
\usepackage{interval}
\DeclareMathOperator{\sinc}{sinc}
\usepackage{lineno}
\usepackage{subfigure}

\journal{journal}

\begin{document}
	
	\begin{frontmatter}
		
		\title{Biologically Inspired Oscillating Activation Functions Can Bridge the Performance Gap between Biological and Artificial Neurons}

		\begin{abstract}
			The recent discovery of special human neocortical pyramidal neurons that can individually learn the XOR function highlights the significant performance gap between biological and artificial neurons. The output of these pyramidal neurons first increases to a maximum with input and then decreases. Artificial neurons with similar characteristics can be designed with oscillating activation functions. Oscillating activation functions have multiple zeros allowing single neurons to have multiple hyper-planes in their decision boundary. This enables even single neurons to learn the XOR function. This paper proposes four new oscillating activation functions inspired by human pyramidal neurons that can also individually learn the XOR function. Oscillating activation functions are non-saturating for all inputs unlike popular activation functions, leading to improved gradient flow and faster convergence. Using oscillating activation functions instead of popular monotonic or non-monotonic single-zero activation functions enables neural networks to train faster and solve classification problems with fewer layers. An extensive comparison of 23 activation functions on CIFAR 10, CIFAR 100, and Imagentte benchmarks is presented and the oscillating activation functions proposed in this paper are shown to outperform all known popular activation functions.
			
		\end{abstract}

		\begin{keyword}
			Oscillating activation function \sep XOR problem \sep neural network \sep convolutional neural network \sep image classification
		\end{keyword}

	\end{frontmatter}
	
	\section{Introduction}
	Artificial neurons are greatly simplified models of biological neurons. Unsurprisingly, a significant performance gap exists between artificial and biological neurons. A particularly severe and fundamental limitation of artificial neurons is that individual neurons can only perform linear classification. Multilayer networks are needed to learn datasets that are not linearly separable. However, the recent discovery of special human neocortical pyramidal neurons that can individually learn the XOR function \cite{doi:10.1126/science.aax6239} suggests that artificial neurons inspired by pyramidal neurons might also be capable of individually learning the XOR function. This paper proposes four new oscillating activation functions inspired by human pyramidal neurons that can individually learn the XOR function and outperform all known popular activation functions on benchmark problems.
	
	A wide variety of functions have been used as activation functions in Artificial Neural Networks (ANNs) since their inception \cite{dubey}. The choice of activation function is a crucial hyperparameter that determines the performance of deep neural networks. Non-linear activation functions endow neural networks with the ability to learn complex non-linear functions. Since the composition of any number of linear functions is also linear, neural networks require non-linear activation functions to solve classification problems that are not linearly separable. Also, it is not known Fa priori if a particular dataset is linearly separable, so non-linear activation functions are generally used. Since ANNs learn by gradient descent in the parameter space, activation functions with different derivative values can significantly change learning speed and final accuracy. 
	
	Saturating activation functions have derivative values arbitrarily close to zero over an interval. Thus we can define non-saturating activation functions as functions with zero derivative values only at isolated points. Since the parameters of an ANN are updated based on the derivative of the loss function with respect to the parameters, saturating activation functions with small derivative values can slow down learning in ANNs. This problem of stagnation of gradient descent due to small derivative values resulting from the use of saturating activation functions becomes worse with increasing number of hidden layers and is referred to as the \emph{vanishing gradient problem} \cite{glorot2010understanding}, \cite{hochreiter1998vanishing}. 
	
	The discovery of the computationally simple Rectified Linear Unit (ReLU) activation that does not saturate for all positive values allowed deeper networks to be trained \citep{Hahnloser}. However, the ReLU activation has zero derivative for all negative inputs and improvements to ReLU like leaky ReLU and Parametric ReLU (PReLU) were introduced. However all ReLU variants are not differentiable at the origin, so a smooth (infinitely differentiable) approximation called Sigmoid Linear Unit (SiLU) was proposed. Both leaky ReLU and SiLU are computationally costlier than the ReLU but provide a performance advantage in many applications \citep{Ramachandran}. 
	
	Universal approximation theorems \citep{Funahashi} guarantee that multilayer networks of sigmoids and ReLU can learn arbitrarily complex continuous functions to any accuracy. Despite the ability of multilayer neural networks to learn arbitrarily complex activation functions, each neuron in a neural network has a single hyperplane as its decision boundary and hence only makes a linear classification \cite{noel2021growing}. Thus single neurons with sigmoidal, ReLU, Swish, and Mish activation functions cannot learn the XOR function. Recent research has discovered that biological neurons in layers two and three of the human cortex have oscillating activation functions and can individually learn the XOR function. The presence of oscillating activation functions in biological neural neurons might partially explain the performance gap between biological and artificial neural networks. This paper proposes four new oscillating activation functions which enable individual neurons to learn the XOR function without manual feature engineering. The paper also explores the possibility of using oscillating activation functions to solve classification problems with fewer neurons and reduce training time. 
	
	At the start of ANN research, the identity function was used as the activation function resulting in a ANNs capable of only solving linear regression and classification problems \cite{nwankpa2018activation}. The output of a single linear neuron or its activation is given by $a = g(z) = z = w^T x+b$. Thus a linear neuron uses an affine function to predict its output from its inputs. Stochastic Gradient Descent (SGD) \cite{bottou2012stochastic} can be used to effectively trains linear neurons. Due to its popularity and rich history, the SGD update rule for a linear neuron is referred to by many well-known names, such as the ADALINE rule, Windrow-Hoff rule, or Delta rule. The linear (identity) activation was inspired by biological neurons that produce linearly increasing output with increasing net input. The composition of any finite number of linear functions is a linear function. Hence, a multilayer network of linear neurons has the same representative power as a single layer of linear neurons. The problem of fitting an affine function to data was well-studied before the advent of neural networks, and networks composed of linear neurons are equivalent to linear regression models. 
	
	The outputs of many types of biological neurons show saturation for inputs beyond a certain threshold; hence sigmoidal neurons were introduced to capture this behavior \cite{apicella2021survey}. Logistic-sigmoidal and tan-sigmoidal neurons capture the saturating nature of biological neurons \cite{nwankpa2018activation}. Tan-sigmoidal neurons are known to outperform Logistic-sigmoidal neurons since logistic-sigmoid always produce positive outputs. These positive outputs can get combined to large positive values saturating the next layer of neurons. In general, activation functions that output symmetrically positive and negative values reduce saturation and alleviate the vanishing gradient problem in deep neural networks \cite{apicella2021survey}.
	
	The simplest model of a biological neuron is the Perceptron model \cite{minsky69perceptrons}. In this model, neurons fire or fail to fire depending on weighted inputs exceeding a threshold. A binary activation function captures this behavior and switches its output between two values (0 and 1 or -1 and 1). Single Perceptrons can be trained efficiently using the Perceptron Learning Algorithm (PLA), which is guaranteed to converge for all linearly separable problems after a finite number of updates using the PLA update rule \cite{minsky69perceptrons}. In the Perceptron model, the Signum function is the activation function. Since the Signum function is discontinuous (non-differentiable) at the origin, the Back Propagation (BP) algorithm cannot be used to train multilayer networks of Perceptron units. 
	
	The output of a Perceptron ($\pm 1$) has the advantage of being interpreted as a Yes/No binary decision. The outputs of neurons can be interpreted as binary decisions only when the activation function produces a bounded output between two values, usually 0 to 1 or -1 to 1. To train multilayer networks with BP while still retaining the ability to interpret outputs as binary decisions, one is forced to consider smooth (differentiable) approximations to the Signum, and Heaviside step functions, namely tan-sigmoid and logistic sigmoid, respectively. Although sigmoidal activations have the advantage of producing an interpretable output, these activations have very small derivatives outside a narrow range leading to the slowing down the training process. For example the tan-sigmoid defined as $\tanh(z)=\frac{exp(z)-exp(-z)}{exp(z)+exp(-z)}$ does not respond to inputs outside the narrow range $[-5 , 5]$ since $exp(-5) < 0.01$. This slowing down of the BP training process in deep networks with saturating activation functions is called the 'vanishing gradient problem and is a serious problem in deep neural networks. The vanishing gradient problem can be alleviated to a large extent by the use of activations that have larger derivatives for a larger range of inputs like ReLU \cite{eckle2019comparison}, leaky ReLU \cite{dubey}, PReLU \cite{HeZR015}, GELU \cite{hendrycks2016gaussian}, Softplus \cite{10.1007/s10489-017-1028-7}, ELU \cite{clevert2016fast}, SiLU \cite{apicella2021survey}, SELU \cite{pedamonti}, Swish \cite{Ramachandran} and Mish \cite{misra} activation functions \cite{apicella2021survey}. These newer activation functions perform better in deep networks, are unbounded, and have larger derivative values \cite{HeZR015}.  
	
	Most activation functions explored previously in neural networks are either monotonic or nearly monotonic, with a single zero at the origin. The following section explores the biological inspiration and mathematical reasons for using oscillatory and highly non-monotonic activation functions with multiple zeros.
	
	\section{Biological motivation for oscillating activation functions}
\begin{figure}[H]
		\center
		\includegraphics[scale=0.25]{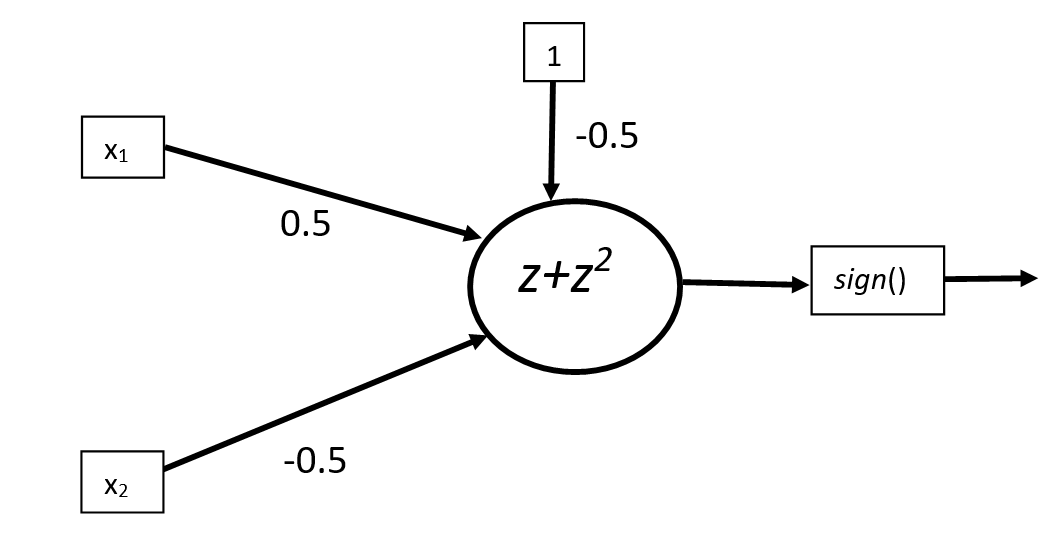}
		\caption{A single neuron solution to the XOR problem using a Shifted Quadratic Unit (SQU).}
		\label{Quadratic_XOR}	
	\end{figure}

\begin{figure}[H]
		\center
		\includegraphics[scale=0.25]{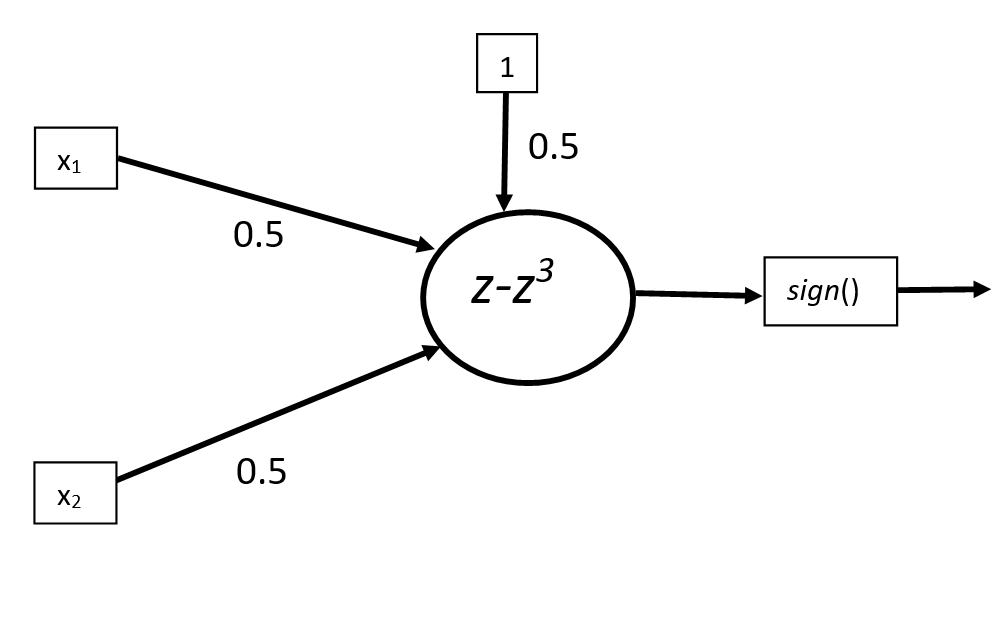}
		\caption{A single neuron solution to the XOR problem using a Non-Monotonic Cubic Unit (NCU).}
		\label{Cubic_XOR}
	\end{figure}

\begin{figure}[H]
		\center
		\includegraphics[scale=0.25]{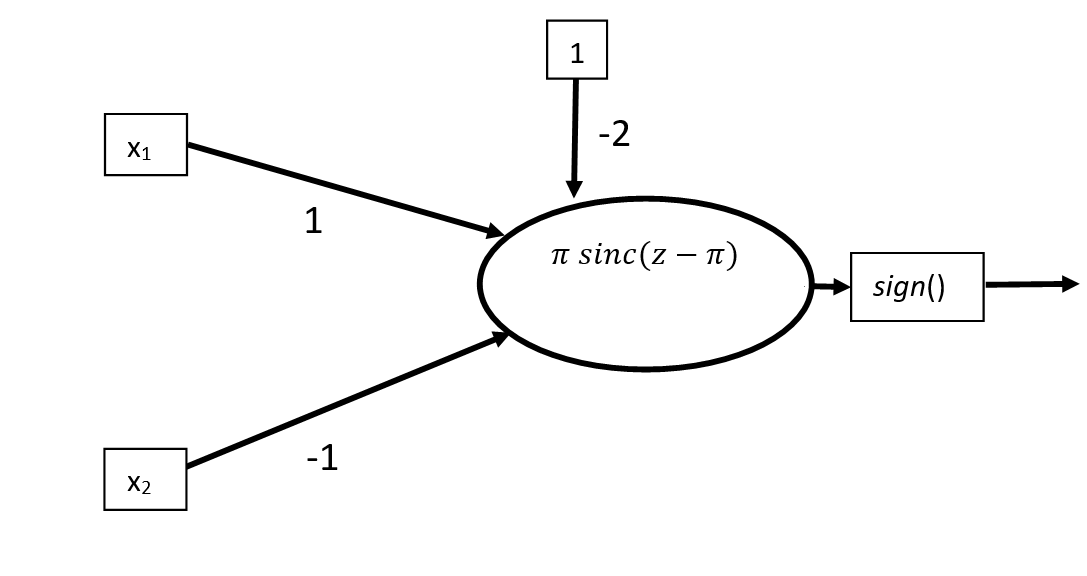}
		\caption{A single neuron solution to the XOR problem using a Shifted Sinc Unit (SSU).}
		\label{sinc_XOR}
	\end{figure}

	Albert Gidon et al., \cite{doi:10.1126/science.aax6239} discovered a new type of pyramidal neuron in the human cerebral cortex that is capable of individually learning the XOR function (a task that is impossible with single artificial neurons using sigmoidal, ReLU, leaky ReLU, PReLU, GELU, Softplus, ELU, SELU, Swish, and Mish activations). A 3-layer neural network with 2 hidden layers and 1 output layer is required to learn the XOR function with all known activation functions. However, the XOR function can be learned by a single neuron with oscillatory activation functions. This is because the decision boundary for a neuron that outputs an activation $a = g(z)=g(w^T z+b)$ is the set of points $z$ for which $g(z) = 0$. If $z=0$ is the only zero of the activation function $g(z)$, then the decision boundary is a single hyperplane $z = w^T z+b = 0$. Two hyperplanes are needed to separate the classes in the XOR dataset, so single neurons with activation functions having a minimum of two zeros can solve the XOR problem. A detailed mathematical analysis of this issue was presented in a previous paper \cite{noel2021growing}. Figures \ref{Quadratic_XOR}, \ref{Cubic_XOR}, and \ref{sinc_XOR} show 3 single neuron solutions to the XOR problem using the new activation functions proposed in this paper. 
	
	Activation is defined to have the "XOR Property" if it can be used in a single neuron to learn the XOR function. Since specific biological neurons responsible for higher-order thinking in the human cerebral cortex have the XOR property, this is also desirable for artificial neurons. \cite{noel2021growing} first presented a single-neuron solution to the XOR problem using the GCU activation function. The XOR problem is the task of learning the following dataset. A bipolar encoding of inputs and outputs is used instead of binary encoding for convenience. 
	
	Fig. \ref{Quadratic_XOR}, Fig. \ref{Cubic_XOR} and Fig. \ref{sinc_XOR} show three new single neuron solutions to the XOR problem. These solutions were obtained by training single neurons on the XOR dataset Eq. \ref{Eq: XOR dataset} using Stochastic Gradient Descent (SGD).
	
	\begin{equation}\label{Eq: XOR dataset}
		D = \left\{ 
		\left(\begin{bmatrix} -1\\-1 \end{bmatrix}, -1\right), \left(\begin{bmatrix} 1\\-1 \end{bmatrix}, 1\right),
		\left(\begin{bmatrix} -1\\1 \end{bmatrix}, 1\right),
		\left(\begin{bmatrix}  1\\1 \end{bmatrix}, -1\right)
		\right\}
	\end{equation}
	
	In the above, the following definition of the $sinc$ function is used:
	\begin{eqnarray}
		\sinc{(z)}&=&\left\{\begin{array}{ll}
			1 &  z= 0;  \\
			\frac{\sin{(z)}}{z} &  \mbox{elsewhere}. \\ 
		\end{array} \right.
	\end{eqnarray}	
	
	This paper presents 4 new oscillating activation functions that allow individual neurons to learn the XOR function and outperform all known activation functions on CIFAR 10, CIFAR 100, and Imagentte benchmarks. The oscillatory activation functions proposed in this paper also outperformed popular activation functions on a wide variety of real-world problems \cite{Golroudbari}, \cite{Bharadwaj}, \cite{Sharma1}, \cite{Sharma2} and \cite{MishraP}. 
	
	\section{Desirable properties of activation functions}
	Although a wide variety of hitherto unexplored non-monotonic and oscillatory activation functions can provide good performance in deep networks, not all functions can serve as the useful activation function for the reasons discussed below. 
	First, useful functions acting as activations must closely approximate the linear (or identity) function for small values. For artificial neurons to mimic biological neurons, they must be linear at the origin. However, there are deep practical reasons for this requirement. The weights of a neuron are initialized to small values at the start of training, so the weighted input $z = w^T x+b$ is close to 0 for all neurons at the start of training. Since the identity function has a derivative of 1 at 0, the neuron has a chance to learn quickly. Conversely, suppose the derivative of activation is small at the origin. In that case, it will cause a slowing down or stalling of the parameter updates after initialization with small values. Thus functions like $\cos{(z)}$, $z^2$, $z^3$  do not perform well as activation functions since these functions have zero derivative at the origin. On the other hand, these poorly performing activation functions can be changed into useful activation functions like $\sin{(z)}$, $z^2+z$, $z-z^3$ that have a derivative of 1 at the origin. \\
			
	\begin{table}[H]
		\centering
		\tiny
		\caption{Definition and properties of activation functions}
		\label{properties}
		\begin{tabular}{|c|c|c|c|c|}
			\hline 
			\multirow{2}{*}{\begin{tabular}{c}
					\textbf{Activation}\\\textbf{Function} \end{tabular}} & \textbf{Equation} & \multirow{2}{*}{\begin{tabular}{c} \textbf{Discontinous} and  \\ \textbf{Non-differentiable points} \end{tabular}}    & \textbf{Monotonicity} & \textbf{Range}                            \\ 
			&&&&\\ \hline
			Signum  &      $f_1(z) = \left\{ \begin{array}{cl}
				-1 &  z < 0; \\
				0 &  z = 0; \\
				1 &  z > 0.
			\end{array}\right.$    & $(z=0)/(z=0)$   & No &  $\left[ -1, 1 \right]$                                                   \\ \hline
			Identity    &  $f_2(z)=z$       & None/None & Yes   &  ($-\infty$,$\infty$)                                                           \\ \hline
			Bipolar Sigmoid   & $f_3(z)=\frac{1-e^{-z}}{1+e^{-z}}$ & None/None  & No &  $\left[-1,1\right]$                                                          \\ \hline
			Sigmoid / Softmax   & $f_4(z)=\frac{1}{1+e^{-z}}$ & None/None     & Yes    & [0,1]                                                          \\ \hline
			Tanh & $f_5(z)=\tanh{(z)}$  & None/None   & Yes                   & $\left[-1,1\right]$                        \\ \hline
			Absolute value   & $f_6(z)=|z|$        &  
			$(z=0)/(z=0)$
			& Yes                 & [0,$\infty$)       \\ \hline
			LiSHT               &      $f_{7}(z)=z\tanh{(z)}$    & None/None    & No                    &  ($-\infty$,$\infty$)          \\ \hline
			Softplus            &   $f_{8}(z)=\ln{(1+e^z)}$       & None/None
			&  Yes    &  [0,$\infty$)   \\ \hline
			ReLU &    $f_{9}(z)=\max{(0,z)}$      & None/$(z=0)$     & Yes    & [0,$\infty$)    \\ \hline
			Leaky ReLU          &  $f_{10}(z)=\left\{\begin{array}{ll}
				0.01z & z<0;  \\
				z & z\geq 0; \\ 
			\end{array} \right.$ &  None/$(z=0)$          & Yes    &  ($-\infty$, $\infty$)         \\ \hline
			GELU&	$f_{11}(z)\approx 0.5z(1+\tanh{(\sqrt{\frac{2}{\pi}}z
				+0.044715z^3)})$
			&  $(z=0)$/None  & Yes    &[-0.5,$\infty$)          \\ \hline
			SELU &   $f_{12}(z)=\left\{\begin{array}{ll}
				\lambda z &  z\geq 0;  \\
				\lambda\alpha (e^z-1) &  z< 0. \\
				\alpha\approx 1.6733 & \lambda \approx 1.0507 
			\end{array} \right.$         & 
			None/$(z=0)$ 
			&     Yes       &  [$-\lambda\alpha$,$\infty$)                             \\ \hline
			Mish                &     $f_{13}(z)=z\tanh{(\ln{(1+e^z)})}$     &         Yes/Yes  &     Yes  & [-0.31,$\infty$)        \\ \hline
			Swish               &  $f_{14}(z)=\frac{z}{1+e^{-z}}$        &  None/None                   & Yes        &  [-0.5,$\infty$)           \\ \hline			
			ELU & $f_{15}(z)=\left\{\begin{array}{ll}
				z &  z\geq 0;  \\
				(e^z-1) &  z< 0. \\ 
			\end{array} \right.$&None/$(z=0)$ &Yes&[-1,$\infty$) \\ \hline
			PReLU &$f_{16}(z)=\left\{\begin{array}{ll}
				z &  z\geq 0;  \\
				\alpha z &  z< 0. \\
				\alpha=0.25 &\\
			\end{array} \right.$&None/$(z=0)$&Yes&($-\infty$,$\infty$) \\ \hline
			Sine Unit (SU)               &  $f_{17}(z)=\sin{(z)}$   & None/None                  & No & $\left[-1,1\right]$     \\ \hline
			\begin{tabular}{c}Shifted \\ Quadratic \\ Unit (SQU) \end{tabular} & $f_{18}(z)=z^2+z$  & None/None   & No                   & $[-0.25,\infty)$                       \\ \hline
			\begin{tabular}{c} Non-Monotonic \\ Cubic  Unit (NCU) \end{tabular} & $f_{19}(z)=z-z^3$  & None/None  & No        & $(-\infty, \infty)$                        \\ \hline
			z2cosz & $f_{20}=z^2\cos{(z)}$ & None/None & No & $(-\infty, \infty) $ \\ \hline
			\begin{tabular}{c} Shifted Sinc \\ Unit (SSU) \end{tabular} & $f_{21}=\pi\sinc{(z-\pi)}$ &  None/None & No & $[-0.68 , \pi ]$  
			\\ \hline
			\begin{tabular}{c} Growing Cosine \\ Unit  (GCU)\end{tabular} & $f_{22}(z) = z\cos{(z)}$  & None/None   & No   & ($-\infty$, $\infty$)            \\ \hline
			\begin{tabular}{c} Decaying Sine \\ Unit (DSU) \end{tabular} & \begin{tabular}{c} $f_{23}(z) = \frac{\pi}{2}(\sinc{(z-\pi)}$ \\ $-\sinc{(z+\pi)})$ \end{tabular} & None/None   & No                   & [-1.04, 1.04]                        \\ \hline
		\end{tabular}
	\end{table}	
	
	A consequence of the linearity for small input values is that the entire network behaves like a linear classifier after initialization with small parameter values. This linearity at initialization has a regularizing effect since the non-linear part of the activation is only used as needed (avoiding overfitting). Another consequence of linearity is that $g(0)=0$, since a linear function has zero value for zero input. \\
	
	Table \ref{properties} and Table \ref{addit} provide a comprehensive summary of different activation functions and their properties. Activation functions that are sign equivalent to the identity function do not have the XOR property as proven in a previous paper \cite{noel2021growing}. A function $f$ is said to be sign-equivalent to a function $g$ iff $sign(f(z))=sign(g(z))$ for all values of $z$. 	Activation functions that are sign equivalent to the identity function are strictly negative for negative inputs, strictly positive for positive inputs and zero only at the origin. \\
	
	Table \ref{addit} shows that all popular activation functions are sign equivalent to the identity function and do not have the XOR property as proved in \cite{noel2021growing}. It is also seen from Table \ref{addit} that all oscillating activation functions proposed in this paper are infinitely differentiable, unbounded and single neurons with these activation functions can learn the XOR function (XOR property) like the pyramidal XOR neurons in the human brain.
	
	\begin{table}[H]
		\centering
		\caption{Properties of Activation Functions}
		\scriptsize
		\label{addit}
		\small
		\begin{tabular}{|c|c|c|c|c|}
			\hline
			\begin{tabular}{c}
				Activation\\
				Function 
			\end{tabular} & \begin{tabular}{c} Small \\ 
				Value\\Approximation \end{tabular} & \begin{tabular}{c} Number \\ of\\hyperplanes\end{tabular}  & \begin{tabular}{c} Sign\\ Equivalence\\ to Identity \end{tabular}  & \begin{tabular}{c} \\XOR \\Property\end{tabular} \\ \hline
			Signum  $(f_1)$           &  NA             &    1       & Yes        &  No \\ \hline
			Identity $(f_2)$           &  $z$          &    1       & Yes       & No    \\ \hline
			Bipolar Sigmoid  $(f_3)$   &  $z$           &  1                     & Yes                             & No                      \\ \hline
			Sigmoid/Softmax   $(f_4)$  &  $z$           &  1                     & Yes                             & No                      \\ \hline
			Tanh       $(f_5)$   &  $z$                         &    1                   &                  Yes            & No                       \\ \hline
			Absolute   $(f_6)$    &   NA                        &  1                     &    No                          &  No                     \\ \hline
			LiSHT     $(f_7)$          &   $z^2$                       & 1                      &  No                            & No                      \\ \hline
			Softplus  $(f_8)$          & $\ln{2}+0.5z$      &    0                    &    No                & No                      \\ \hline
			ReLU      $(f_9)$          &    NA                       &   1                    &  No                            &    No                   \\ \hline
			Leaky ReLU    $(f_{10})$      &   NA                         &  1                     & Yes                             &     No                  \\ \hline
			GELU         $(f_{11})$      &    $0.5z$                       &  1  & Yes                             & No                       \\ \hline
			SELU        $(f_{12})$        &     NA                      &   1                    &  Yes                            &   No                    \\ \hline
			Mish        $(f_{13})$        &    $ z\tanh{(\ln{2})} $                       &  1                     & Yes                             &   No    \\ \hline
			Swish   $(f_{14})$            &    $0.5z$                       &  1            & Yes                             &    No                   \\ \hline
			ELU $(f_{15})$ & $z$ & 1 &  Yes & No \\ \hline
			PReLU $(f_{16})$ & NA & 1 &  Yes & No \\ \hline 			
			SU    $(f_{17})$        &   $z$                        &  $\infty$  & No                              &   Yes                    \\ \hline
			SQU   $(f_{18})$           &    $z$                      &  2                     & No                             &   Yes                    \\ \hline
			NCU     $(f_{19})$           &    $z$                       &  3                     & No                             &   Yes                    \\ \hline
			z2cosz $(f_{20})$ & $z^2$ & $\infty$ &  No & Yes  \\ \hline
			SSU $(f_{21})$ & $z$ & $\infty$ & No & yes \\ \hline 
			GCU $(f_{22})$      &    $z$       &  $\infty$                     & No                             &   Yes                    \\ \hline
			DSU  $(f_{23})$              &    $z$                      &  $\infty$                     & No                             &   Yes                    \\ \hline	
		\end{tabular}
	\end{table}

\begin{figure}[H]
		\centering
		\tiny
		\subfigure[]{
			\includegraphics[scale=0.4]{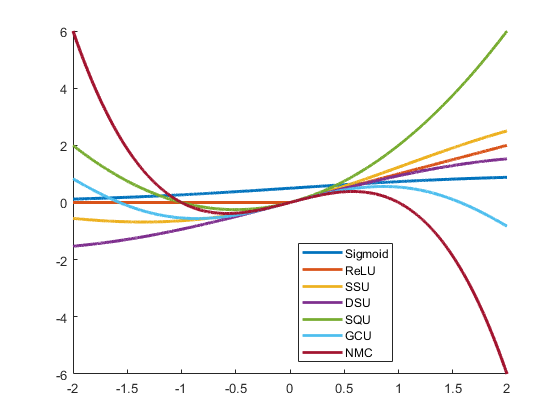}}
		\subfigure[]{
			\includegraphics[scale=0.4]{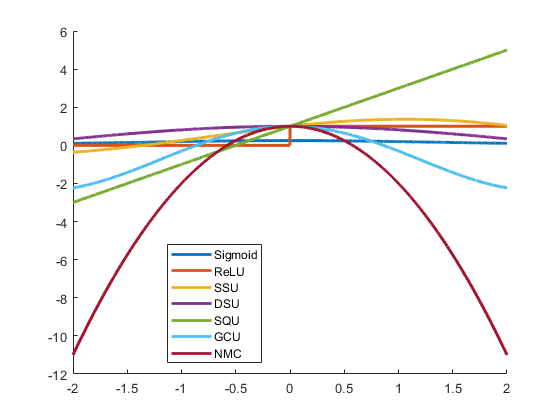}}
		\caption{Plot of (a) activation function and (b) derivatives.} 
		\label{activations}
	\end{figure}
	
Fig. \ref{activations} shows plots of different activation function and their derivatives. Fig. \ref{activations} limits the plots to the domain $[-2 , 2]$ to avoid scaling issues that arise from differences in range between different activation functions. It is seen that all oscillating activation functions are linear close to the origin but differ drastically from popular sigmoidal and ReLU like activation functions away from the origin. In particular oscillating activation functions cross the X-axis at multiple points and assume positive as well as negative values for both positive and negative inputs. It is also clear from Fig. \ref{activations} that oscillating activation functions like NMC have significantly larger derivative values compared to popular activation functions which explains the faster learning with backpropagation (Fig. \ref{cifar10}, Fig. \ref{cifar100} and Fig. \ref{fig:Imagenette}).
	
\section{Results}
	The performance of oscillating activation functions proposed in this paper was compared to all known popular activation functions on CIFAR 10, CIFAR 100, and Imagentte benchmarks \cite{krizhevsky2009learning}. A comprehensive set of 23 activation functions were taken for comparison to determine the best activation function. Figure \ref{model} in Appendix 1 provides the CNN architectures used with different benchmarks. The best-known hyperparameter settings were used for each benchmark. The sparse categorical cross-entropy loss function was used since all benchmarks considered are multi-class classification problems. Table \ref{hyper} below provides detailed hyperparameter settings. The average accuracy over 5 independent training experiments was considered to reduce the effects of possible lucky or unlucky random weight initialization.
	
	\begin{table}[tbh!]
		\centering
		\caption{Network and hyper-parameter settings}
		\label{hyper}
		\begin{tabular}{|c|c|c|c|c|c|}  \hline
			Dataset & Optimizer &Batch Size& Learning Rate & Decay & $EP_{max}$  \\ \hline
			Imagenette & RMSprop&64 & $10^{-5}$ & None & 25 \\ \hline
			CIFAR10 & Adam&32 & $10^{-4}$&$10^{-5}$& 40 \\ \hline
			CIFAR100 & Adam&32 &  $10^{-4}$&$10^{-5}$& 40 \\ \hline  
		\end{tabular}
	\end{table}

	Tables \ref{cifar10}, \ref{cifar100}, and \ref{imagenette} shows the difference in performance when different activation functions are used to solve the CIFAR-10, CIFAT-100, and Imagenette benchmark problem. From Table \ref{cifar10}, it is seen that oscillating activation functions occupy 4 out of the top 5 positions on the CIFAR-10 benchmark. On CIFAR-100 oscillating activation functions occupy all top 5 positions (\ref{cifar100}). On the Imagenette benchmark, the most complex benchmark used in this study, oscillating activation functions occupy 4 out of the top 5 positions. Overall the oscillating activation functions occupy the top 2 positions in all benchmarks.

	\begin{table}[H]
		\scriptsize
		\centering
		\caption{CIFAR10 Results}
		\label{cifar10}
		\begin{tabular}{|c|c|c|c|c|c|}
			\hline
			\begin{tabular}{c}
				Activation\\
				Function 
			\end{tabular} & \begin{tabular}{c} Mean Test\\ 
				Accuracy \end{tabular} & \begin{tabular}{c} STD Test\\ Accuracy\end{tabular}  & \begin{tabular}{c} Mean Test\\Loss\end{tabular}  & \begin{tabular}{c}STD Test\\Loss\end{tabular} &\begin{tabular}{c}Performance\\Rank\end{tabular}\\ \hline
			Signum  $(f_1)$ &0.4247600019&0.0193138897&	1.6367679358&0.0617457247&23 \\ \hline
			Identity $(f_2)$ &0.7696999907&0.0029826172&0.7296342969&0.0026068858&14 \\ \hline
			Bipolar Sigmoid $(f_3)$&0.7502400041&0.0025523318&0.7395075321&0.0078517050&18\\ \hline
			Sigmoid/Softmax  $(f_4)$&0.5642400026&0.0157303062&1.2281183004&0.0396219308&22\\ \hline
			Tanh       $(f_5)$&0.7805600047&0.0022791258&0.6617408514&0.0082593364&11 \\ \hline
			Absolute   $(f_6)$&0.7609000087&0.0068070504&0.6818503618&0.0194255227&17\\ \hline
			LiSHT     $(f_7)$&0.7037199855&0.0137080096&0.8573125720&0.0393833626&19\\ \hline
			Softplus  $(f_8)$&0.6485200047&0.0063167745&1.0251185656&0.0147744921&21 \\ \hline
			ReLU      $(f_9)$&0.7907400131&0.0037648896&0.6239251137&0.0099365174&6 \\ \hline
			Leaky ReLU    $(f_{10})$&0.7900600076&0.0026340843&0.6231829524&0.0090547343&7 \\ \hline
			GELU  $(f_{11})$&0.7642000079&0.0020059858&0.7242264748&0.0026891309&16\\ \hline
			SELU        $(f_{12})$&0.7732000113&0.0020513537&0.7120848298&0.0091205381&13 \\ \hline
			Mish        $(f_{13})$&0.7776600003&0.0027015418&0.6875398874&0.0096158528&12 \\ \hline
			Swish   $(f_{14})$&0.7653200030&0.0047321826&0.7265463471&0.0056851930&15 \\ \hline
			ELU $(f_{15})$ &0.7872399926&0.0036412093&0.6726237774&0.0090619936&9 \\ \hline
			PReLU $(f_{16})$&0.7943599939&0.0044777513&0.6301926136&0.0114728166&3\\ \hline 			
			SU    $(f_{17})$&0.7829599857&0.0034354656&0.6506057143&0.0091237685&10  \\ \hline
			SQU   $(f_{18})$ &0.8012200117&0.0026678854&0.6182821035&0.0030939807&1 \\ \hline
			NCU     $(f_{19})$ &0.7931800008&0.0024701297&0.6304274797&0.0050465081&4\\ \hline
			z2cosz $(f_{20})$ &0.6564599991&0.0334281753&0.9829714179&0.0874925022&20 \\ \hline
			SSU $(f_{21})$ &0.7949000001&0.0034479072&0.6174252748&0.0094551736&2\\ \hline 
			GCU $(f_{22})$  &0.7916200042&0.0014344330&0.6240387678&0.0045099729&5\\ \hline
			DSU  $(f_{23})$ &0.7898800015&0.0027102726&0.6225146890&0.0073510728&8  \\ \hline	
		\end{tabular}
	\end{table}
	
	\begin{table}[H]
		\scriptsize
		\centering
		\caption{CIFAR100 Results}
		\label{cifar100}
		\begin{tabular}{|c|c|c|c|c|c|}
			\hline
			\begin{tabular}{c}
				Activation\\
				Function 
			\end{tabular} & \begin{tabular}{c} Mean Test\\ 
				Accuracy \end{tabular} & \begin{tabular}{c} STD Test\\ Accuracy\end{tabular}  & \begin{tabular}{c} Mean Test\\Loss\end{tabular}  & \begin{tabular}{c}STD Test\\Loss\end{tabular}& \begin{tabular}{c}Performance\\Rank\end{tabular} \\ \hline
			Signum  $(f_1)$ &0.1631599993&0.0059768197&3.6133363247&	0.0300070019
			& 22\\ \hline
			Identity $(f_2)$ &0.4537200034&	0.0021395364&2.1713591099&0.0066851386 	&12 \\ \hline
			Bipolar Sigmoid $(f_3)$ & 0.4148800075&0.0035067915&2.3039673328&0.0078262696&17\\ \hline
			Sigmoid/Softmax  $(f_4)$  &0.0487200007&0.0774400018&4.3591092587&0.4921222210&23
			\\ \hline
			Tanh       $(f_5)$ &0.4739799917&0.0030029313&2.0305715561&0.0096993330&8 \\ \hline
			Absolute   $(f_6)$ &0.3987000048&0.0039181660&2.3327541351&0.0259482760&18	\\ \hline
			LiSHT     $(f_7)$ &0.3893799961&0.0058659742&2.4357059002&0.0283739347&19 \\ \hline
			Softplus  $(f_8)$ &0.3241400003&0.0057735977&2.7978433132&0.0232604334&21  \\ \hline
			ReLU      $(f_9)$ &0.4450199902&0.0081499438&2.1725291252&0.0359708507&14 \\ \hline
			Leaky ReLU    $(f_{10})$ &0.4515199959&0.0037365229&2.1373793125&0.0129646756&13 \\ \hline
			GELU  $(f_{11})$ &0.4408399999&0.0031135780&2.2239931583&0.0128022986&15\\ \hline
			SELU        $(f_{12})$&0.4631200016&0.0020855688&2.1128221512&0.0122153134&10 \\ \hline
			Mish        $(f_{13})$ &0.4553400040&0.0045797771&2.1502437115&0.0112833609& 11\\ \hline
			Swish   $(f_{14})$&0.4375400007&0.0028938555&2.2452141285&0.0143031591&16 \\ \hline
			ELU $(f_{15})$ &0.4815799952&0.0021451311&2.0192164898&0.0125542093& 7\\ \hline
			PReLU $(f_{16})$ &0.4680999994&	0.0037035140&2.0796844959&0.0213182969&9 \\ \hline 			
			SU    $(f_{17})$ &0.4869200051&	0.0023215516&	1.9846775293&	0.0043500624&6
			\\ \hline
			SQU   $(f_{18})$ &0.4945400000&	0.0047906512&1.9733317137&	0.0099660246&2
			\\ \hline
			NCU     $(f_{19})$ &0.4921400011&0.0055134777&1.9592674255&	0.0124952843&3
			\\ \hline
			z2cosz $(f_{20})$ &0.3692399979&0.0091482508&2.5151033878&	0.0439377521&20
			\\ \hline
			SSU $(f_{21})$ &0.4880400062&0.0045071492&1.9598821402&	0.0161290765&5
			\\ \hline 
			GCU $(f_{22})$  &0.4962600052&0.0049029027&1.9300193787&0.0140280468&1
			\\ \hline
			DSU  $(f_{23})$ &0.4917800009&	0.0041777546&	1.9397407293&	0.0203836128&4
			\\ \hline	
		\end{tabular}
	\end{table}
	
	\begin{table}[H]
		\scriptsize
		\caption{Imagenette Results}
		\label{imagenette}
		\begin{tabular}{|c|c|c|c|c|c|}
			\hline
			\begin{tabular}{c}
				Activation\\
				Function 
			\end{tabular} & \begin{tabular}{c} Mean Test\\ 
				Accuracy \end{tabular} & \begin{tabular}{c} STD Test\\ Accuracy\end{tabular}  & \begin{tabular}{c} Mean Test\\Loss\end{tabular}  & \begin{tabular}{c}STD Test\\Loss\end{tabular}&  \begin{tabular}{c}Performance\\Rank\end{tabular} \\ \hline
			Signum  $(f_1)$ & 0.1043566883	&0.0018694451&	2.7776308060	&0.0133106500&20\\ \hline
			Identity $(f_2)$ & 0.5950573206&0.0151592069&	2.5428726673	&0.1907586862 & 11 \\ \hline
			Bipolar Sigmoid $(f_3)$  & 0.4218089163&	0.0082472304&1.6774385214&0.0104293088&18
			\\ \hline
			Sigmoid/Softmax  $(f_4)$  &0.1013503179&0.0057793409&2.3072936058&	0.0024763653&21\\ \hline
			Tanh      $(f_5)$   & 0.6602802515&0.0045706764&1.2063075066&0.0316241230&4	  \\ \hline
			Absolute   $(f_6)$  &0.5558726072&0.0146326491&1.3625491381&0.0424726152&14\\ \hline
			LiSHT     $(f_7)$ &0.0909554139&0.0000000000&2.3026307583&0.0000010070 &22  \\ \hline
			Softplus  $(f_8)$ &0.2219108224&	0.1025836351&2.1023106337&0.1714400325&19  \\ \hline
			ReLU      $(f_9)$ &0.5852738857&0.0189149658&1.2827849865&0.0594888064&12 \\ \hline
			Leaky ReLU    $(f_{10})$ & 0.5800764203&0.0118971866&1.3054905891&0.0291386250&13 \\ \hline
			GELU  $(f_{11})$ &0.4411719680&0.0051879458&1.6778397322&0.0198294917&17\\ \hline
			SELU        $(f_{12})$&0.6555414081&0.0036317938&1.7032670259&0.0500926845&6  \\ \hline
			Mish        $(f_{13})$ &0.4512101948&0.0060506173&1.7278023481&0.0629856614&15 \\ \hline
			Swish   $(f_{14})$&0.4408662379&0.0020547128&1.6483911753&0.0096471617&16 \\ \hline
			ELU $(f_{15})$ &0.6528917313&0.0055693046&1.4622555971&0.0294364658&7 \\ \hline
			PReLU $(f_{16})$ &0.6038726091&0.0060213722&1.2444277287&0.0169353213&10\\ \hline 			
			SU    $(f_{17})$ &0.6593121052&0.0140653601&1.2748115301&0.0710113069&5 \\ \hline
			SQU   $(f_{18})$ &0.6143184781&0.0059279277&3.9766886711&0.3287157923&9\\ \hline
			NCU     $(f_{19})$ & 0.6925859928&	0.0076119779&1.0458305478&0.0368917701&1\\ \hline
			z2cosz $(f_{20})$ &0.0909554139	&0.0000000000&2.3026299477&0.0000007979&22 \\ \hline
			SSU $(f_{21})$ &0.6421401381&0.0073974932&1.1889729261&0.0097117699&8\\ \hline 
			GCU $(f_{22})$  &0.6807643294&0.0064474363&1.1340459108&0.0228075988&2\\ \hline
			DSU  $(f_{23})$ &0.6656305790&0.0059415034	&1.0325865984&0.0134966556&3 \\ \hline	
		\end{tabular}
	\end{table}

\begin{figure}[H]
		\centering
		\tiny
		\subfigure[]{
			\includegraphics[scale=0.25]{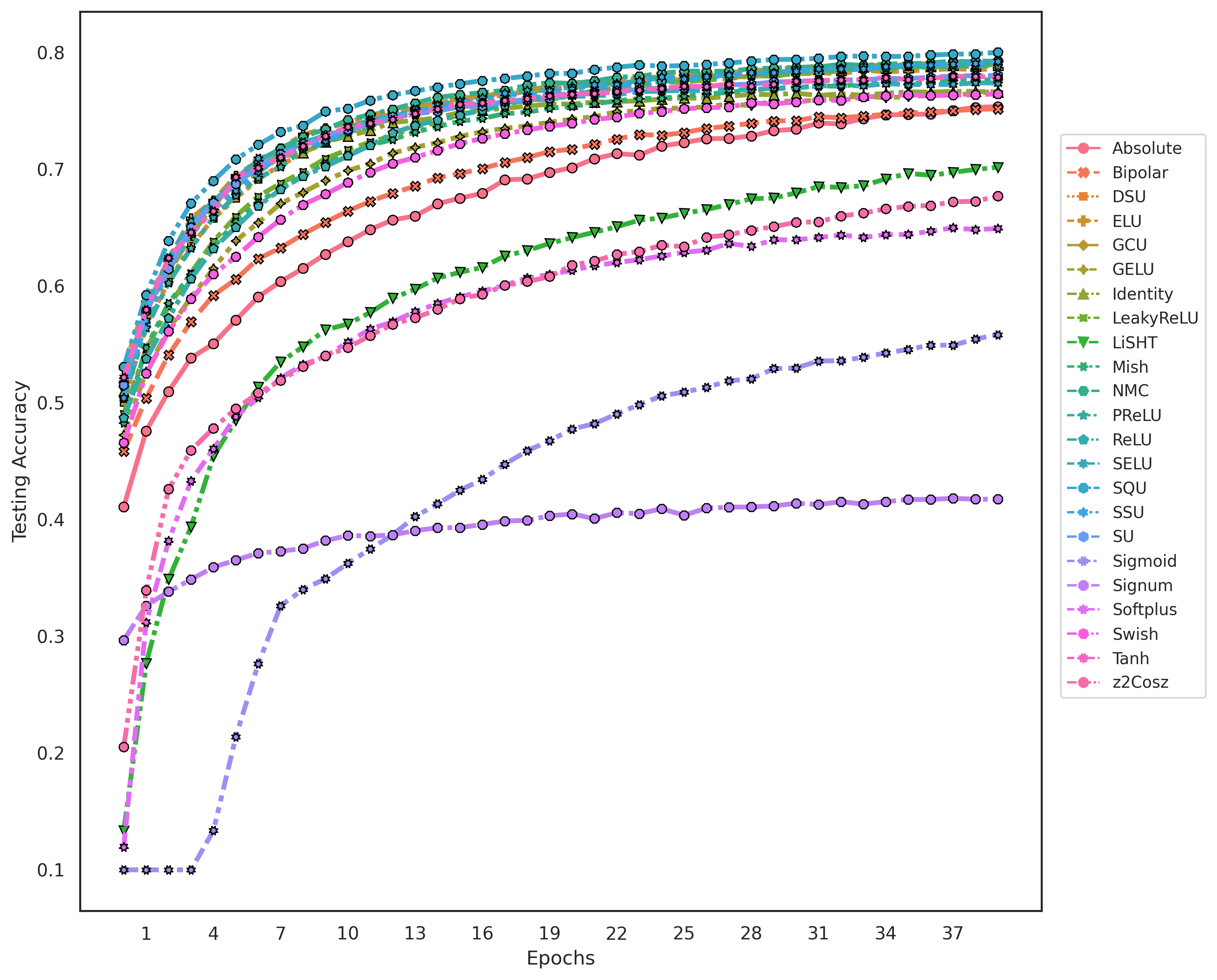}}
		\subfigure[]{
			\includegraphics[scale=0.25]{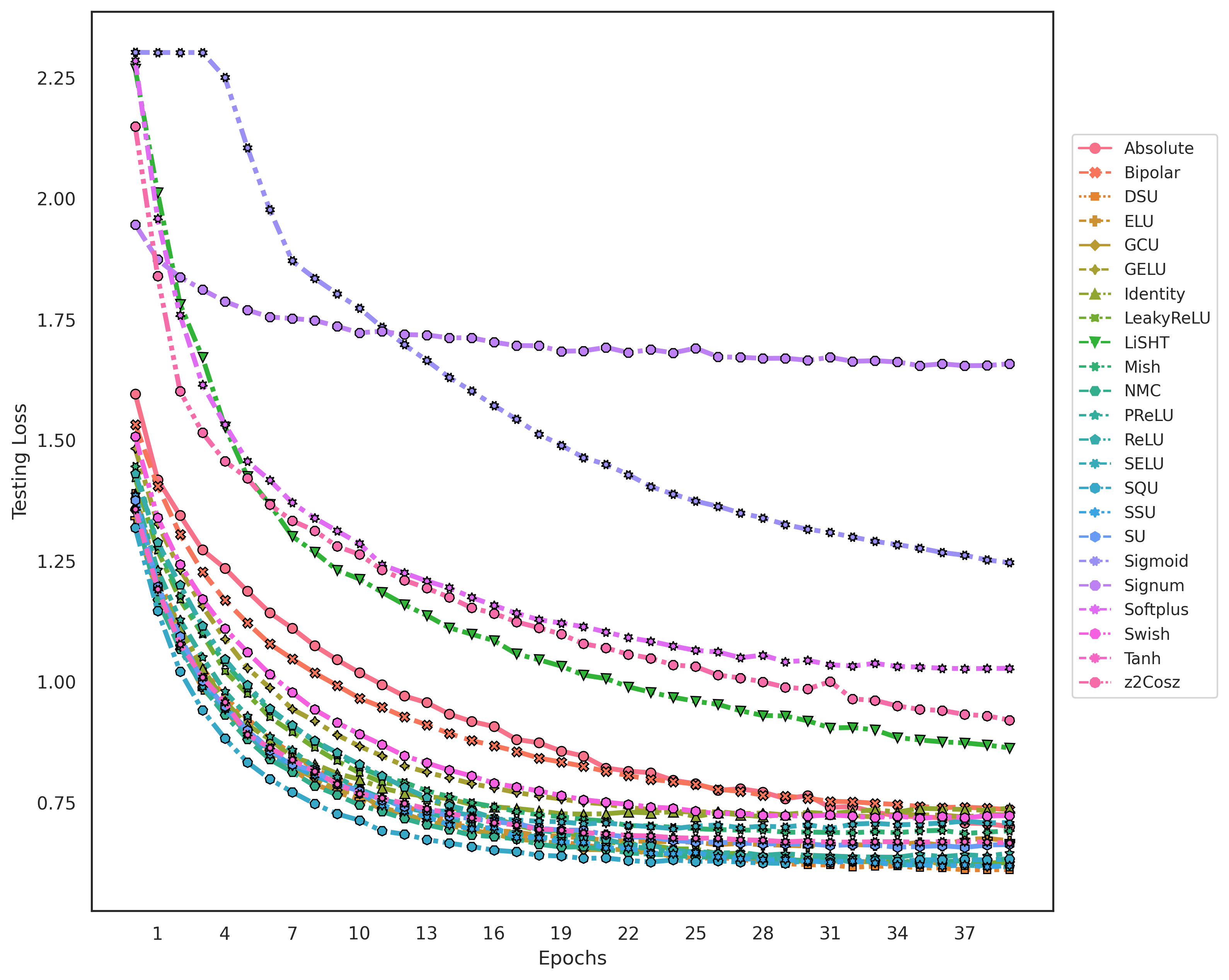}}
		\caption{Training performance on CIFAR-10.}
		\label{fig:CIFAR10}
	\end{figure}

\begin{figure}[H]
		\centering
		\tiny
		\subfigure[]{
			\includegraphics[scale=0.25]{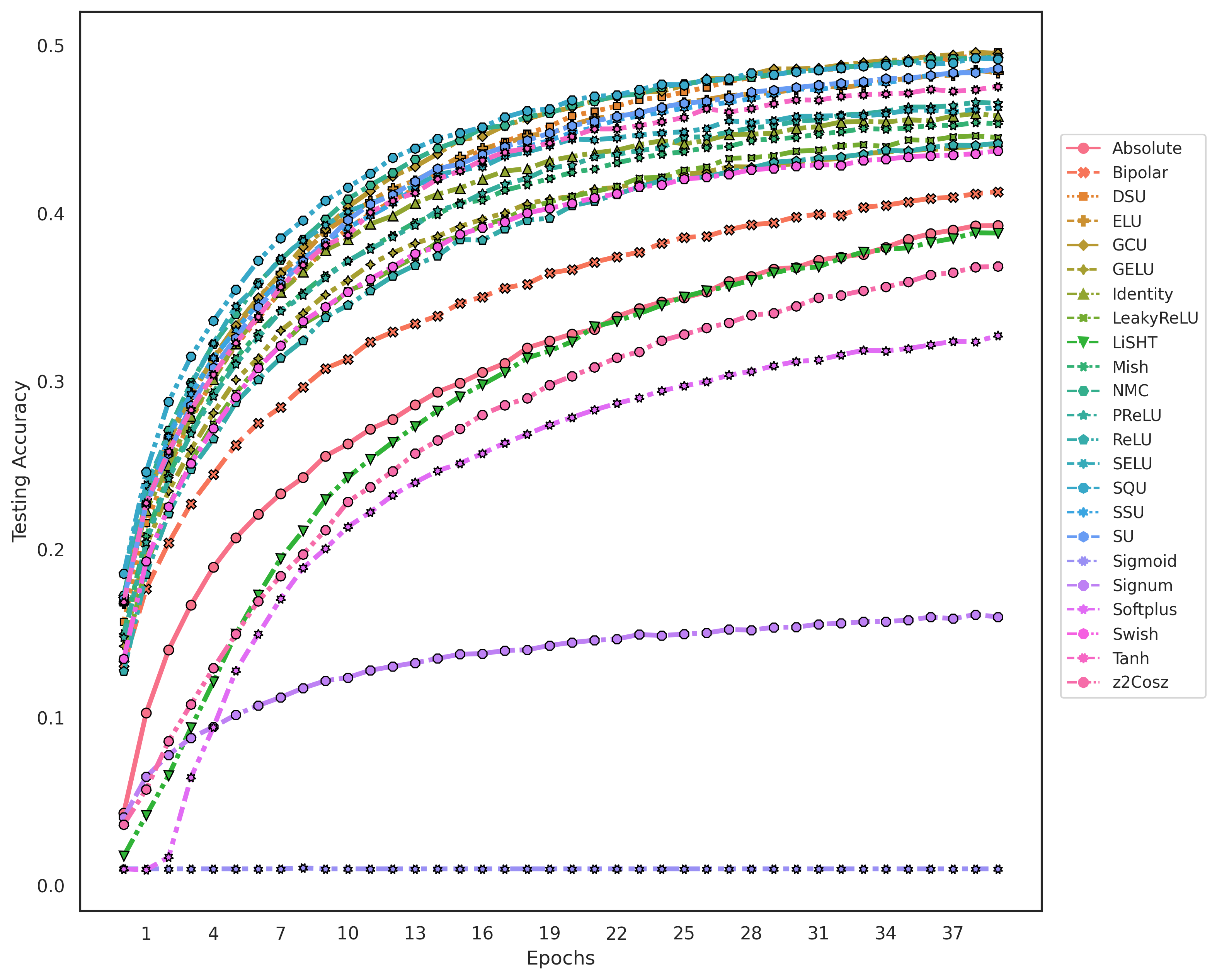}}
		\subfigure[]{
			\includegraphics[scale=0.25]{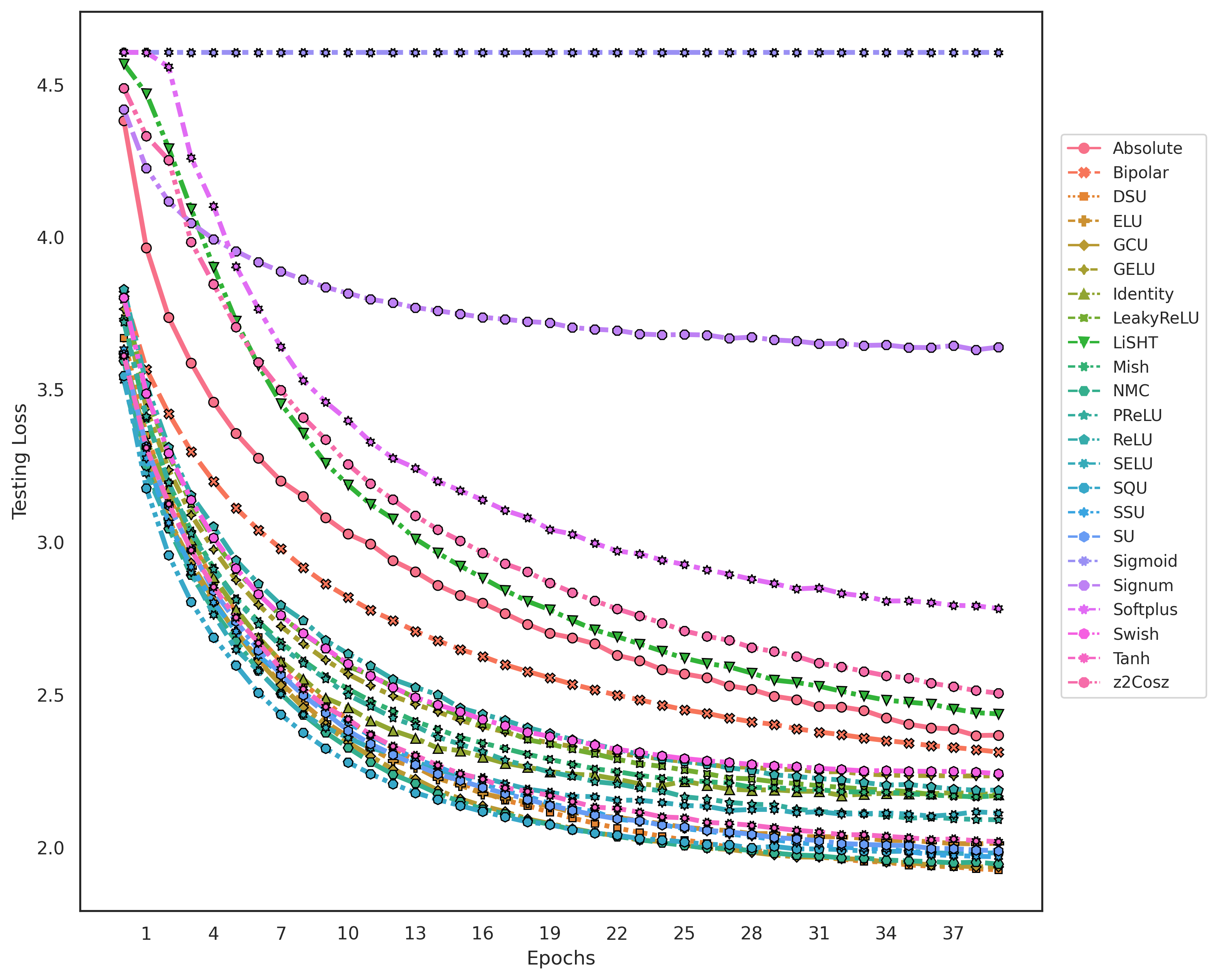}}
		\caption{Training performance on CIFAR-100.}
		\label{fig:CIFAR100}
	\end{figure}

\begin{figure}[H]
		\centering
		\tiny
		\subfigure[]{
			\includegraphics[scale=0.25]{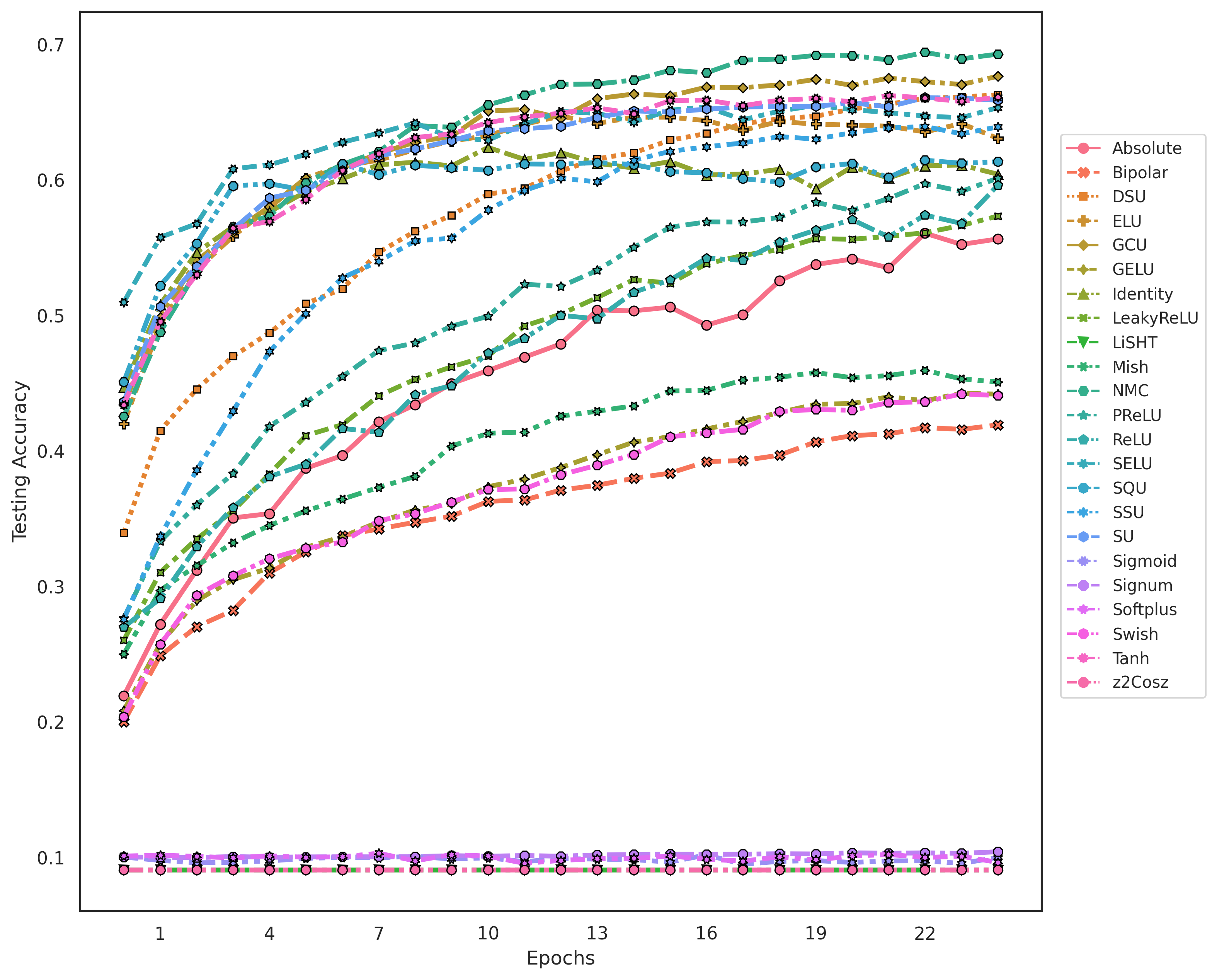}}
		\subfigure[]{
			\includegraphics[scale=0.25]{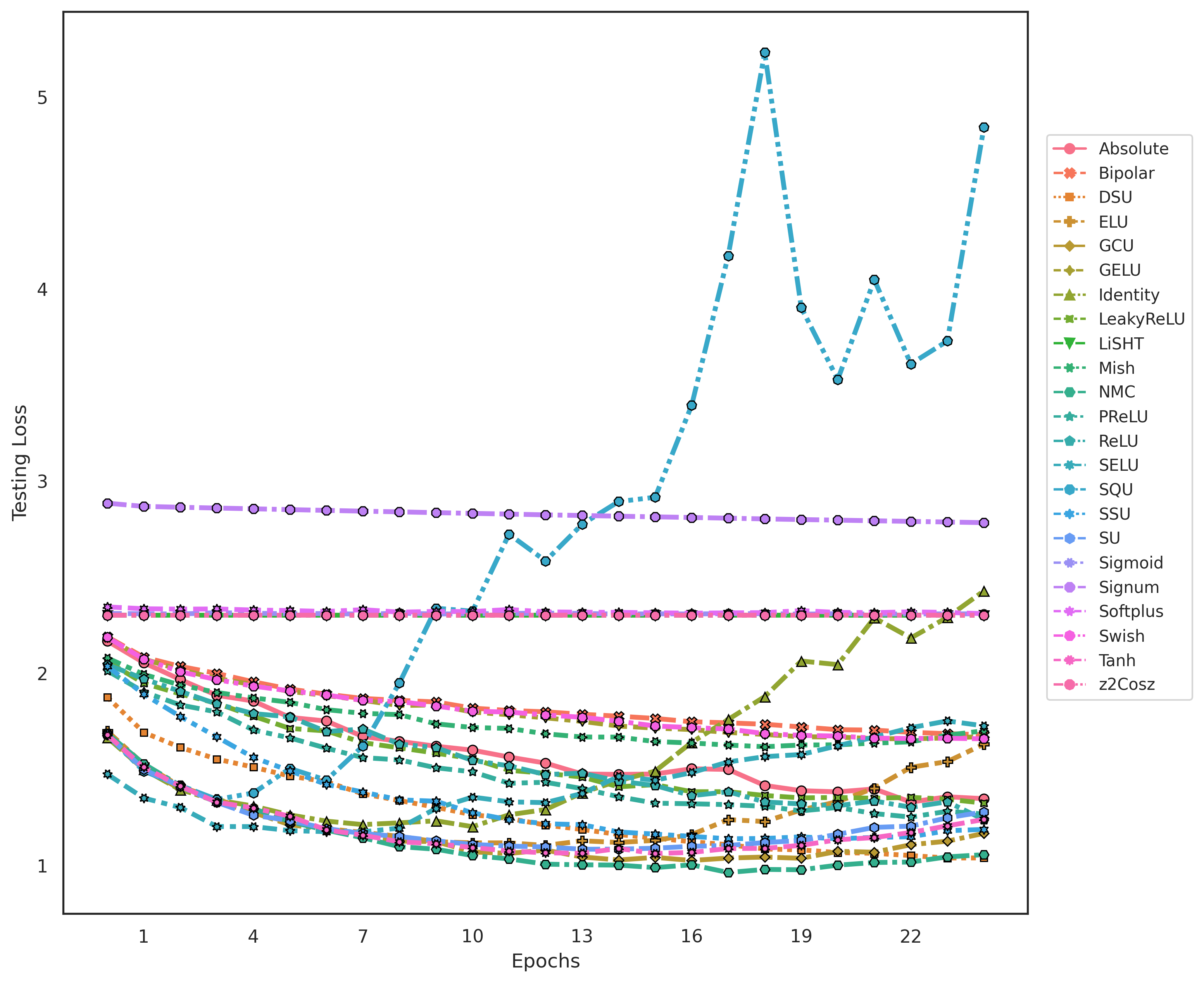}}
		\caption{Training performance on Imagenette.}
		\label{fig:Imagenette}
	\end{figure}

Table \ref{summary} compares the overall performance of all 23 activation functions based on the average rank on all benchmarks. It is seen that oscillating activation functions occupy the top-5 positions. The activation function $z^2\cos(z)$ has zero derivatives at the origin and performs poorly as expected (Section 1.2). Fig. \ref{fig:CIFAR10}, Fig. \ref{fig:CIFAR100} and Fig. \ref{fig:Imagenette} show the training performance on CIFAR-10, CIFAR-100 and Imagenette benchmarks. These results indicate that networks with oscillating activation functions consistently train faster.

\begin{table}[H]
		\centering
		\caption{Summary of Results}
		\label{summary}
		\begin{tabular}{||c|c|c|c|c|c||}
			\hline
			Function  & CIFAR10 & CIFAR100 &	ImageNette& A.R. & Overall Position \\ \hline 
			Signum  &	23 &22& 20&	21.667&	22 \\
			Identity &	14&	12& 11&	12.333&	13 \\
			Biploar  &	18 &17&  18&	17.667&	18  \\
			Sigmoid  & 22 &23&  21 &22.000 &23 \\
			Tanh      &	11 &8&  4 & 7.667 &8  \\
			Absolute  	&17 &	18&  14 &	16.333&	17 \\
			LiSHT   &	19 & 19&  22 &	20.000&	19\\
			Softplus &	21&	21 &  19 &20.333 &20 \\
			ReLU &	6 &	14&  12 &10.667 &11  \\
			LeakyReLU  &7 &	13& 13	&11.000 &	12 \\
			GELU  &16 &15 &  17 &16.000	&16  \\
			SELU  &	13&	10&	 6&	9.667&	10 \\
			Mish  &	12&	11&	15&	12.667&	14  \\
			Swish &15 &16&  16&	15.667&	15 \\
			ELU   &	9&	7&  7&	7.667&	8 \\
			PReLU 	&3&	9& 10&	7.333&	7 \\
			SU     &	10 &6& 5&	7.000&	6\\
			SQU    &1&	2&  9&	4.000&	3  \\
			NMC   	&4&	3& 1&	2.667&	1 \\
			z2Cosz &20&	20 & 22&20.667&	21 \\
			SSU     &2 &	5& 8 &	5.000 &	4 \\
			GCU    	&5&	1& 2 &	2.667 &	1 \\
			DSU    &	8 &	4 & 3	&5.000&	4 \\ \hline
		\end{tabular}
	\end{table}
	
	\vspace{2cm}
	
	\vspace{2cm}
	
	\begin{figure}[H]
		\center
		\includegraphics[scale=0.7]{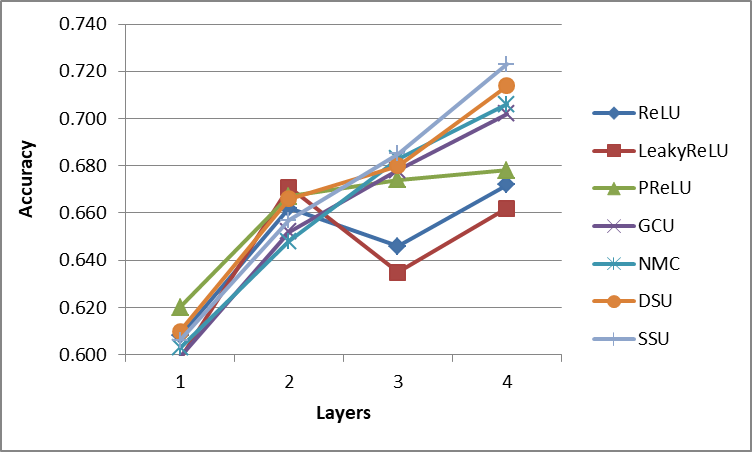}
		\caption{Variation of accuracy with the number of layers on the CIFAR-10 benchmark.}
		\label{layers}	
	\end{figure}
	
	Fig. \ref{layers} shows the increase in accuracy with the increasing number of convolutional layers. It is clearly seen that oscillating activation functions significantly improve accuracy over popular activation functions when the number of convolutional layers is increased. Stated differently, higher accuracy is achieved with the same number of layers. This is to be expected since each neuron with an oscillating activation function can partition its input space with more than one hyperplane. Since neurons with oscillating activation functions have multiple hyperplanes in the decision boundary, fewer neurons are needed in classification tasks. This higher representative power of neurons with oscillating activation function was also demonstrated with single neuron solutions to the XOR problem (Fig. \ref{Quadratic_XOR}, Fig. \ref{Cubic_XOR} and Fig.  \ref{sinc_XOR}).

\section{Conclusion}   
	The discovery of single neurons in the human brain with oscillating activation functions capable of individually learning the XOR function is a biological inspiration for a new class of oscillating activation functions. Oscillating activation functions have multiple hyperplanes in their decision boundary, enabling neurons to make more complex decisions than popular sigmoidal, ReLU-like Swish, and Mish activation functions. The higher representative power of networks with oscillating activation functions allows classification and regression tasks to be solved with fewer neurons. Also, the oscillations in the activation function appear to improve gradient flow and speed up backpropagation learning. This paper proposed 4 new biologically inspired oscillating activations Shifted Quadratic Unit (SQU), Non-Monotonic Cubic (NCU), Shifted Sinc Unit (SSU), and Decaying Sine Unit (DSU)) that enable single neurons to learn the XOR problem. The new activation functions proposed in this paper outperform all popular activation functions on the  CIFAR-10, CIFAR-100, and Imagenette benchmarks. Oscillating activation functions proposed in this paper also provide a performance advantage on many real-world problems \cite{Golroudbari} - \cite{Sharma2}. More importantly, oscillating activation functions are shown to allow ANNs to learn successfully with fewer neurons since each neuron can divide its input space with multiple hyperplanes. The results presented in this paper suggest that deep networks with oscillating activation functions might partially bridge the performance gap between biological and artificial neural networks.
	
	
	\section*{Appendix 1}
	
	\begin{figure}[H]
		\centering
		\tiny
		\subfigure[]{
			\includegraphics[scale=0.18]{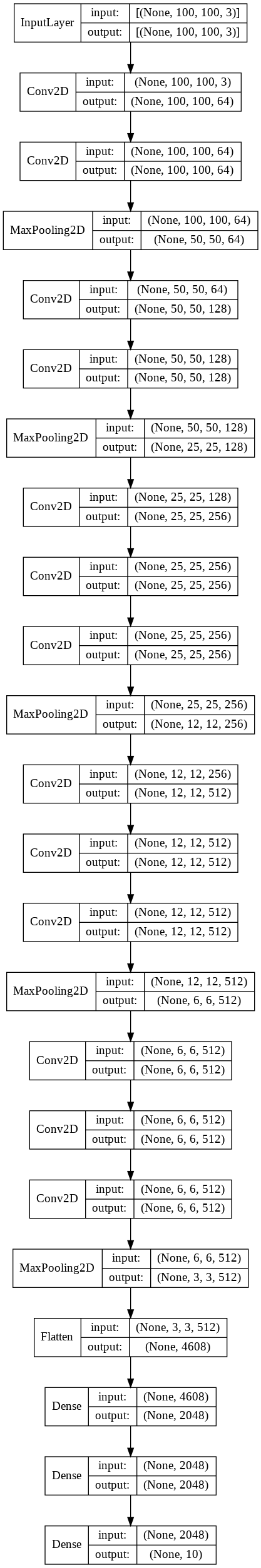}}
		\hspace{3cm}
		\subfigure[]{
			\includegraphics[scale=0.20]{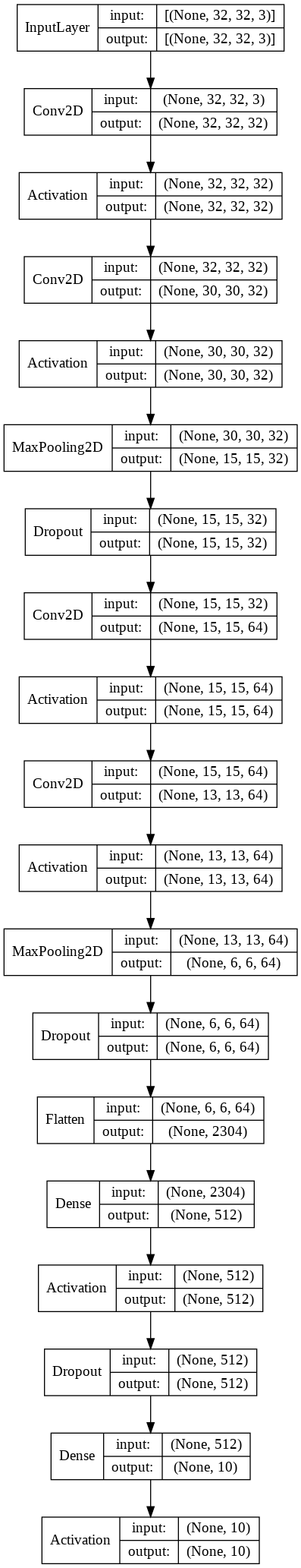}}
		\label{model}
		\caption{Model architectures for (a)ImageNette and (b) CIFAR10 \& CIFAR100}
	\end{figure}
	
	\bibliographystyle{plain}

\end{document}